\documentclass[11pt,a4paper]{article}
\usepackage[hyperref]{acl2020}
\usepackage{times}
\usepackage{latexsym}
\usepackage{graphicx}

\usepackage{ulem}
\usepackage{enumitem,kantlipsum}

\usepackage{microtype}

\aclfinalcopy % Uncomment this line for the final submission

\title{Experimental Evaluation and Development of a Silver-Standard for the {MIMIC}-{III} Clinical Coding Dataset}

\author{Thomas Searle$^1$, Zina Ibrahim$^1$, Richard JB Dobson$^{1,2}$ \\
  $^1$Department of Biostatistics and Health Informatics,\\
  Institute of Psychiatry, Psychology and Neuroscience,\\
  King’s College London, London, U.K.\\
  $^2$Institute of Health Informatics, University College London,\\
  London, London, U.K.\\
    \{firstname.lastname\}@kcl.ac.uk}
\date{}

\begin{document}
\maketitle
\begin{abstract}
Clinical coding is currently a labour-intensive, error-prone, but critical administrative process whereby hospital patient episodes are manually assigned codes by qualified staff from large, standardised taxonomic hierarchies of codes. Automating clinical coding has a long history in NLP research and has recently seen novel developments setting new state of the art results. A popular dataset used in this task is MIMIC-III, a large intensive care database that includes clinical free text notes and associated codes. We argue for the reconsideration of the validity MIMIC-III's assigned codes that are often treated as gold-standard, especially when MIMIC-III has not undergone secondary validation. This work presents an open-source, reproducible experimental methodology for assessing the validity of codes derived from EHR discharge summaries. We exemplify the methodology with MIMIC-III discharge summaries and show the most frequently assigned codes in MIMIC-III are under-coded up to 35\%. 
\end{abstract}

\section{Introduction}
Clinical coding is the process of translating statements written by clinicians in natural language to describe a patient's complaint, problem, diagnosis and treatment, into an internationally-recognised coded format \cite{World_Health_Organisation2011-eo}. Coding is an integral component of healthcare and provides standardised means for reimbursement, care administration, and for enabling epidemiological studies using electronic health record (EHR) data \cite{Henderson2006-ma}. 

Manual clinical coding is a complex, labour-intensive, and specialised process. It is also error-prone due to the subtleties and ambiguities common in clinical text and often strict timelines imposed on coding encounters. The annual cost of clinical coding is estimated to be \$25 billion in the US alone \cite{Farkas2008-yn}.
 
To alleviate the burden of the status quo of manual coding, several Machine learning (ML) automated coding models have been developed \cite{Larkey1996-ce,Aronson2007-tb,Farkas2008-yn,Perotte2014-ce,Ayyar2016-sn,Baumel2018-gp,Mullenbach2018-oq,Falis2019-hg}. However, despite continued interest, translation of ML systems into real-world deployments has been limited. An important factor contributing to the limited translation is the fluctuating quality of the manually-coded real hospital data used to train and evaluate such systems, where large margins of error are a direct consequence of the difficulty and error-prone nature of manual coding. To our knowledge, the literature contains only two systematic evaluations of the quality of clinically-coded data, both based on UK trusts and showing accuracy to range between 50 to 98\% \citet{Burns2012-pm} and error rates between 1\%-45.8\% \citet{CHKS_Ltd2014-kl} respectively. In \citet{Burns2012-pm}, the actual accuracy is likely to be lower because the reviewed trusts used varying statistical evaluation methods, validation sources (clinical text vs clinical registries), sampling modes for accuracy estimation (random vs non-random), and the quality of validators (qualified clinical coders vs lay people). \citet{CHKS_Ltd2014-kl} highlight that 48\% of the reviewed trusts used discharge summaries alone or as the primary source for coding an encounter, to minimise the amount of raw text used for code assignment. However, further portions of the documented encounter are often needed to assign codes accurately. 

The Medical Information Mart for Intensive Care (MIMIC-III) database \cite{Johnson2016-np} is the largest free resource of hospital data and constitutes a substantial portion of the training of automated coding models. Nevertheless, MIMIC-III is significantly under-coded for specific conditions \cite{Kokotailo2005-ic}, and has been shown to exhibit reproducibility issues in the problem of mortality prediction \cite{Johnson2017-pr}. Therefore, serious consideration is needed when using MIMIC-III to train automated coding solutions.

In this work, we seek to understand the limitations of using MIMIC-III to train automated coding systems. To our knowledge, no work has attempted to validate the MIMIC-III clinical coding dataset for all admissions and codes, due to the time-consuming and costly nature of the endeavour. To illustrate the burden, having two clinical coders, working 38 hours a week re-coding all 52,726 admission notes at a rate of 5 minutes and \$3 per document, would amount to $\sim$\$316,000 and $\sim$115 weeks work for a `gold standard' dataset. Even then, documents with a low inter-annotator agreement would undergo a final coding round by a third coder, further raising the approximate cost to $\sim$\$316,000 and stretching the 70 weeks.

In this work, we present an experimental evaluation of coding coverage in the MIMIC-III discharge summaries. The evaluation uses text extraction rules and a validated biomedical named entity recognition and linking (NER+L) tool, MedCAT \cite{Kraljevic2019-tj} to extract ICD-9 codes, reconciling them with those already assigned in MIMIC-III. The training and experimental setup yield a reproducible open-source procedure for building silver-standard coding datasets from clinical notes. Using the approach, we produce a silver-standard dataset for ICD-9 coding based on MIMIC-III discharge summaries.

This paper is structured as follows: Section \ref{sec:background} reviews essential background and related work in automated clinical coding, with a particular focus on MIMIC-III. Section \ref{sec:setup} presents our experimental setup and the semi-supervised development of a silver standard dataset of clinical codes derived from unstructured EHR data. The results are presented in Section \ref{sec:results}, while Section \ref{sec:discussion} discusses the wider impact of the results and future work. 

\section{Background}\label{sec:background}

\subsection{Clinical Coding Overview}
The International Statistical Classification of Diseases and Health Related Problems (ICD) provides a hierarchical taxonomic structure of clinical terminology to classify morbidity data \cite{World_Health_Organisation2011-eo}. The framework provides consistent definitions across global health care services to describe adverse health events including illness, injury and disability. Broadly, patient encounters with health services result in a set of clinical codes that directly correlate to the care provided. 

Top-level ICD codes represent the highest level of the hierarchy, with ICD-9/10 (ICD-10 being the later version) listing 19 and 21 chapters respectively. Clinically meaningful hierarchical subdivisions of each chapter provide further specialisation of a given condition. 

Coding clinical text results in the assignment of a single primary diagnosis and further secondary diagnosis codes \cite{World_Health_Organisation2011-eo}. The complexity of coding encounters largely stems from the substantial number of available codes. For example, ICD-10-CM is the US-specific extension to the standard ICD-10 and includes ~72,000 codes. Although a significant portion of the hierarchy corresponds to rare conditions, `common' conditions to code are still in the order of thousands. 

Moreover, clinical text often contains specialist terminology, spelling mistakes, implicit mentions, abbreviations and bespoke grammatical rules. However, even qualified clinical coders are not permitted to infer codes that are not explicitly mentioned within the text. For example, a diagnostic test result that indicates a condition (with the condition not explicitly written), or a diagnosis that is written as `questioned' or `possible' cannot be coded. 

Another factor contributing to the laborious nature of coding is the large amount of duplication present in EHRs, as a result of features such as copy \& paste being made available to clinical staff. It has been reported that 20-78\% of clinicians duplicate sections of records between notes \cite{Bowman2013-xx}, subsequently producing an average data redundancy of 75\% \cite{Zhang2017-ya}.

\subsection{ MIMIC-III - a Clinical Coding Database}\label{sec:mimic_3}
MIMIC-III \cite{Johnson2016-np} is a  de-identified database containing data from the intensive care unit of the Beth Israel Medical Deaconess Center, Boston, Massachusetts, USA, collected 2001-12. MIMIC-III is the world's largest resource of freely-accessible hospital data and contains demographics, laboratory test results, procedures, medications, caregiver notes, imaging reports, admission and discharge summaries, as well as mortality (both in and out of the hospital) data of 52,726 critical care patients. MIMIC provides an open-source platform for researchers to work on real patient data. At the time of writing, MIMIC-III has over 900 citations. 

\subsection{Automated Clinical Coding}\label{sec:general_work}
Early ML work on automated clinical coding  considered ensembles of simple text classifiers to predict codes from discharge summaries \cite{Larkey1996-ce}. Rule-based models have also been formulated, by directly replicating coding manuals. A prominent example of rule-based models is the BioNLP 2007 shared task \cite{Aronson2007-tb}, which supplied a gold standard labelled dataset of radiology reports. The dataset continues to be used to train and validate ML coding. For example, \citet{Kavuluru2015-lh} used the dataset in addition to two US-based hospital EHRs. Although the two additional  datasets used by \citet{Kavuluru2015-lh} were not validated to a gold standard, they are reflective of the diversity found in clinical text. Their largest dataset contained  71,463 records, 60,238 distinct code combinations and had an average document length of 5303 words. 

The majority of automated coding systems are trained and tested Using MIMIC-III. \citet{Perotte2014-ce} trained hierarchical support vector machine models on the MIMIC-II EHR \cite{Saeed2011-gf}, the earlier version of MIMIC. The models were trained using the full ICD-9-CM terminology, creating baseline results for subsequent models of 0.395 F1-micro score. \citet{Ayyar2016-sn} used a long-short-term-memory (LSTM) neural network to predict ICD-9 codes in MIMIC-III. However, \citet{Ayyar2016-sn} cannot be directly compared to former methods as the model only predicts the top nineteen level codes. 

Methodological developments continued to use MIMIC-III with Tree-of-sequence LSTMs \cite{Xie2018-br}, hierarchical attention gated recurrent unit (HA-GRU) neural networks \cite{Baumel2018-gp} and convolutional neural networks with attention (CAML) \cite{Mullenbach2018-oq}. The HA-GRU and CAML models were directly compared with \cite{Perotte2014-ce}, achieving 0.405 and 0.539 F1-micro respectively. A recent empirical evaluation of ICD-9 coding methods predicted the top fifty ICD-9 codes from MIMIC-III, suggesting condensed memory networks as a superior network topology \cite{Huang2018-ag}. 

\section{Semi-Supervised Extraction of Clinical Codes} \label{sec:setup}
In this section, we describe the data preprocessing, methodology and experimental design for evaluating the coding quality of MIMIC-III discharge summaries. We also describe the semi-supervised creation of a silver-standard dataset of clinical codes from unstructured EHR text based on MIMIC-III discharge summaries.

\subsection{Data Preparation}\label{sec:data_prep}
Discharge summary reports are used to provide an overview for the given hospital episode. Automated coding systems often only use discharge reports as they contain the salient diagnostic text \cite{Perotte2014-ce,Baumel2018-gp,Mullenbach2018-oq} without over burdening the model. MIMIC-III discharge summaries are categorised distinctly from other clinical text. The text is often structured with section headings and content section delimiters such as line breaks. We identify Discharge Diagnosis (DD) sections in the majority of discharge summary reports 92\% (n=48,898) using a simple rule based approach. These sections are lists of diagnoses assigned to the patient during admission. \citet{Xie2018-br} previously used these sections to develop a matching algorithm from discharge diagnosis to ICD code descriptions with moderate success demonstrating state-of-the-art sensitivity (0.29) and specificity (0.33) scores. For the 8\% (n=3,828) that are missing these sections we manually inspect a handful of examples and observe instances of patient death and administration errors. The SQL procedures used to extract the raw data from a locally built replica of the MIMIC-III database and the extraction logic for DDs are available open-source as part of this wider analysis\footnote{https://tinyurl.com/t7dxn3j}. 

Table \ref{tab:extracted_diagnosis} lists example extracted DDs. There is a large variation in structure, use of abbreviations and extensive use of clinical terms. Some DDs list the primary diagnosis alongside secondary diagnosis, whereas others simply list a set of conditions. 

\begin{table}[]
    \centering
    \begin{tabular}{|p{4.5cm}|p{1.5cm}|}
        \hline
        \textbf{Extracted Discharge Diagnosis} & \textbf{Admission ID} \\
        \hline
        CAD now s/p CABG
        \newline HTN, DM, Osteoarthritis, Dyslipidemia
        & 102894 \\
        \hline
        Left convexity, tentorial, parafalcine Subdural hematoma & 161919 \\
        \hline
        Primary Diagnoses:
        \newline 1. Acute ST segment Elevation Myocardial Infarction
        \newline Secondary Diagnoses:
        \newline 1. Hypertension
        \newline 2. Hyperlipidemia & 152382 \\
        \hline
        Seizures. & 132065 \\
        \hline
    \end{tabular}
    \caption{Example discharge diagnosis subsections extracted from MIMIC-III discharge summaries}
    \label{tab:extracted_diagnosis}
\end{table}
 
\subsection{Semi-Supervised Named Entity Recognition and Linkage Tool}\label{sec:clean_method}
We use MedCAT \cite{Kraljevic2019-tj}, a pre-trained named entity recognition and linking (NER+L) model, to identify and extract the corresponding ICD codes in a discharge summary note. MedCAT utilises a fast dictionary-based algorithm for direct text matches and a shallow neural network concept to learn fixed length distributed semantic vectors for ambiguous text spans. The method is conceptually similar to Word2Vec \cite{Mikolov2013-ib} in that word representations are learnt by detecting correlations of context words, and learnt vectors exhibit the semantics of the underlying words. The tool can be trained in a unsupervised or a  supervised manner. However, unlike Word2Vec that learns a single representation for each word, MedCAT enables the learning of `concept' representations by accommodating synonymous terms, abbreviations or alternative spellings.

We use a MedCAT model pre-loaded with the Unified Medical Language System \cite{Bodenreider2004-ir} (UMLS). UMLS is a meta-thesaurus of medical ontologies that provides rich synonym lists that can be used for recognition and disambiguation of concepts. Mappings from UMLS to the ICD-9\footnote{https://bioportal.bioontology.org/ontologies/ICD9CM} taxonomy are then used to extract UMLS concept to ICD codes. Our large pre-trained MedCAT UMLS model contains $\sim$1.6 million concepts. This model cannot be made publicly available due to constraints on the UMLS license, but can be trained in an an unsupervised method in $\sim$1 week on MIMIC-III with standard CPU only hardware\footnote{https://tinyurl.com/yadtnz3w}.

In an effort to keep our analysis tractable we limit our MedCAT model to only extract the 400 ICD-9 codes that occur most frequently in the dataset. This equates to 76\% (n=48,2379) of total assigned codes (n=634,709). We exclude the other 6,441 codes that occur less frequently. Future work could consider including more of these codes.

\subsection{Code Prediction Datasets}\label{sec:code_prediction_datasets}
We run our MedCAT model over each extracted DD subsection. The model assigns each token or sequence of tokens a UMLS code and therefore an associated ICD code. In our comparison of the MedCAT produced annotations with the MIMIC-III assigned codes we have 3 distinct datasets:

\begin{enumerate}[leftmargin=*]
    \item MedCAT does not identify a concept and the code has been assigned in MIMIC-III. Denoted \textbf{A\_NP} for `Assigned, Not Predicted'.
    \item MedCAT identifies a concept and this matches with an assigned code in MIMIC-III. Denoted \textbf{P\_A} for `Predicted, Assigned'.
    \item MedCAT identifies a concept and this does \emph{not} match with an assigned code in MIMIC-III dataset. Denoted \textbf{P\_NA} for `Predicted, Not Assigned'.
\end{enumerate}

We do not consider the case where both MedCAT and the existing MIMIC-III assigned codes have missed an assignable code as this would involve manual validation of all notes, and as previously discussed, is infeasible for a dataset of this size. 

\subsubsection{Producing the Silver Standard}
Given the above initial datasets we produce our final silver-standard clinical coding dataset by:
\begin{enumerate}[leftmargin=*]
    \item Sampling from the missing predictions dataset (A\_NP) to manually collect annotations where our out-of-the-box MedCAT model fails to recognise diagnoses.
    \item Fine-tuning our MedCAT model with the collected annotations and re-running on the entire DD subsection dataset producing updated A\_NP, P\_A, P\_NA datasets.
    \item Sampling from P\_NA and P\_A and annotating predicted diagnoses to validate correctness of the MedCAT predicted codes.
    \item Exclusion of any codes that fail manual validation step as they are not trustworthy predictions made by MedCAT.
\end{enumerate}

We use the MedCATTrainer annotator \cite{Searle2019-fd} to both collect annotations (stage 1) and to validate predictions from MedCAT (stage 3). To collect annotations, we manually inspect 10 randomly sampled predictions for each of the 400 unique codes from A\_NP and add further acronyms, abbreviations, synonyms etc for diagnoses if they are present in the DD subsection to improve the underlying MedCAT model. To validate predictions from P\_A and P\_NA, we use the MedCATTrainer annotator to inspect 10 randomly sampled predictions for each of the 179 \& 182 unique codes respectively found. We mark each prediction made by MedCAT as correct or incorrect and report results in Section \ref{sec:results_validation}.

\section{Results}\label{sec:results}
The following section presents the distribution of manually collected annotations from sampling A\_NP, our validation of updated P\_A and P\_NA post MedCAT fine-tuning, and the final distribution of codes found in our produced silver standard dataset.

Adding annotations to selected text spans directly adds the spans to the MedCAT dictionary, thereby ensuring further text spans of the same content are annotated by the model - if the text span is unique. We collect 864 annotations after reviewing 4000 randomly sampled DD notes from the A\_NP (Assigned, Not Predicted) dataset. 21.6\% of DDs provide further annotations suggesting that the majority of missed codes lie outside the DD subsection, or are incorrectly assigned.

\begin{figure}
    \centering
    \includegraphics[scale=0.33]{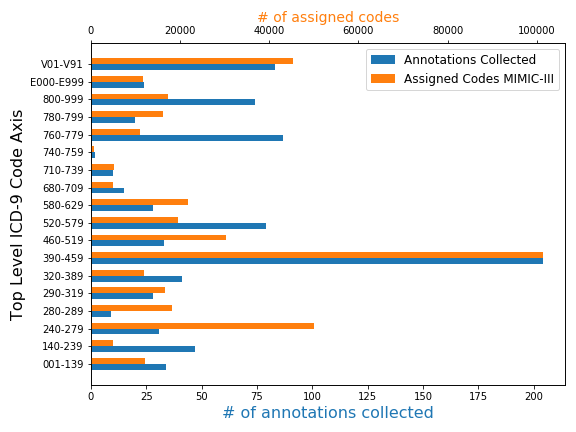}
    \caption{The distributions of manually annotated ICD-9 codes and the assigned codes in MIMIC-III grouped by top-level ICD-9 axis.}
    \label{fig:manually_collected_annotation_distribution}
\end{figure}

Figure \ref{fig:manually_collected_annotation_distribution} shows the distributions of manually collected code annotations and the current MIMIC-III set of clinical codes, grouped by their top-level axis as specified by ICD-9-CM hierarchy. 

We collect proportionally consistent annotations for most groups, including the 390-459 chapter (Diseases Of The Circulatory System), which is the top occurring group in both scales. However, for groups such as 240-279 (endocrine, nutritional and metabolic diseases) and 460-519 (diseases of the respiratory system) we see proportionally fewer manually collected examples despite the high number of occurrence of codes assigned within MIMIC-III. We explain this by the DD subsection lacking appropriate detail to assign the specific code. For example codes under 250.* for diabetes mellitus and the various forms of complications are assigned frequently but often lack the appropriate level of detail specifying the type, control status and the manifestation of complication. 

Using the manual amendments made on the 864 new annotations, we re-run the MedCAT model on the entire DD subsection dataset, producing updated P\_NA, P\_A and A\_NP datasets. We acknowledge A\_NP likely still includes cases of abbreviations, synonyms as we only subsampled 10 documents per code allowing for further improvements to the model.

The MedCAT fine-tuning process was run until convergence as measured by precision, recall and F1 achieving scores 0.90, 0.92 and 0.91 respectively on a held out a test-set with train/test splits 80/20. The fine-tuning code is made available\footnote{https://github.com/tomolopolis/MIMIC-III-Discharge-Diagnosis-Analysis/blob/master/Run\_MedCAT.ipynb}. Annotations are available upon request given the appropriate MIMIC-III licenses.

\subsection{P\_A \& P\_NA Validation}\label{sec:results_validation}
We use the MedCATTrainer interface to validate our MedCAT model predictions in the `Predicted, Assigned' (P\_A) and `Predicted, Not Assigned' (P\_NA) datasets. We sample (a maximum of) 10 unique predictions for each ICD-code resulting in 179 \& 182 ICD-9 codes and 1588 \& 1580 manually validated predictions from P\_A and P\_NA respectively. The validation of code assignment is performed by a medical informatics PhD student with no professional clinical coding experience and a qualified clinical coder, marking each term as correct or incorrect. We achieve good agreement with a Cohen's Kappa of 0.85 and 0.8 resulting in 95.51\% and 87.91\% marked correct for P\_A and P\_NA respectively. We exclude from further experiments all codes that fail this validation step as they are not trustworthy predictions made by MedCAT.

\subsection{Aggregate Assigned Codes \& Codes Silver Standard }
We proportionally predict $\sim$10\% (n=42,837) of total assigned codes (n=432,770). We predict $\sim$16\% of total assigned codes (n=258,953) if we only consider the 182 codes that resulted in at least one matched assignment to those present in the MIMIC-III assigned codes. 

We label and gather our three datasets into a single table, with an extra column called `validated', with values: `yes' for codes that have matched with an assigned code (P\_A), `new\_code' for newly discovered codes (P\_NA), and `no' for codes that we were not able to validate (A\_NP). We have made this silver-standard dataset available alongside our analysis code\footnote{https://tinyurl.com/u8yae8n}.

\subsection{Undercoding in MIMIC-III}
This work aims to identify inconsistencies and variability in coding accuracy in the current MIMIC-III dataset. Ultimately to rigorously identify undercoding of clinical text full, double blind manual coding would be performed. However, as previously discussed, this is prohibitively expensive. 

Comparing the codes predicted by MedCAT to the existing assigned codes enables the development of an understanding of specific groups of codes that exhibit possible undercoding. In this section we firstly show the effectiveness of our method in terms of DD subsection prediction coverage. We then present our predicted code distributions against the MIMIC-III assigned codes at the ICD code chapter level, highlighting the most prevalent missing codes and showing correlations between document length and prevalence.

\subsubsection{Prediction Coverage}\label{sec:pred_coverage}
\begin{figure}
    \centering
    \includegraphics[scale=0.45]{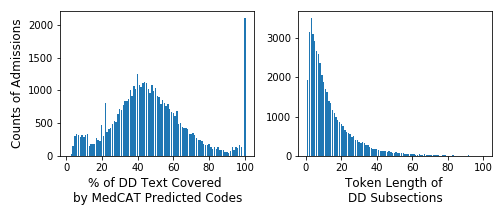}
    \caption{Left: Counts of admissions and the associated \% of characters covered by MedCAT code predictions. Right:Distribution of DD token lengths}
    \label{fig:dd_pred_cov}
\end{figure}

MedCAT provides predictions at the text span level, with only one concept prediction per span. We can therefore calculate the breadth of coverage of our predictions across all DD subsections. Figure \ref{fig:dd_pred_cov} shows the proportion of DD subsection text that are included in code predictions. We note the 100\% proportion (n=2105) is 75\% larger than the next largest indicating that we are often utilising the entire DD subsection to suggest codes although the majority of the coverage distribution is around the 40-50\% range.

We find a token length distribution of DD subsections with $\mu=$14.54, $\sigma=$15.9, $Med=10$ and $IQR=14$ and a code extraction distribution with $\mu=3.6$ and $\sigma=3.1$, $Med=3$ and $IQR=4$ suggesting the DD subsections are complex and often list multiple conditions of which we identify, on average, 3 to 4 conditions.

\subsubsection{Predicted \& Assigned}\label{sec:pred_assned}
\begin{figure}
    \centering
    \includegraphics[scale=.39]{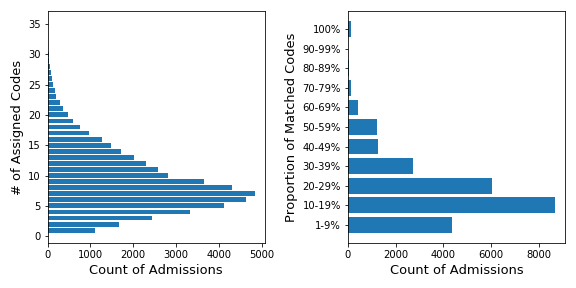}
    \caption{Proportions of matching predictions against total  number of assigned codes per admission.}
    \label{fig:distribution_of_matched_compared_with_assigned_codes}
\end{figure}

Figure \ref{fig:distribution_of_matched_compared_with_assigned_codes} shows the distributions of the number of assigned codes and the proportion of matches grouped into buckets of 10\% intervals. We see a high proportion of matches in assigned codes in the 1-40\% range, indicating that although the DD subsection does contribute to the assigned ICD codes, many of the assigned codes are still missed. We exclude the admissions that had 0 matched codes and discuss this result further in Section \ref{sec:a_np}. 

If we order codes by the number of predicted and assigned we find the three highest occurring codes (4019, 41404, 4280) in MIMIC-III also rank highest in our predictions. However, we note that these three common codes only yield 25-39\% of their total assigned occurrence, which could be explained by these chronic conditions not being listed in the DD subsection and referred elsewhere in the note. If we normalise predictions by their prevalence, we are most successful in matching specific conditions applicable to preterm newborns (7470, 7766), pulmonary embolism (41519) and liver cancer (1550), all of which we match between 69-55\% but rank 114-305 in total prevalence. We suggest these diagnoses are either acute, or the primary cause of an ICU admission so will be specified in the DD subsection.

\begin{figure}
    \centering
    \includegraphics[scale=.34]{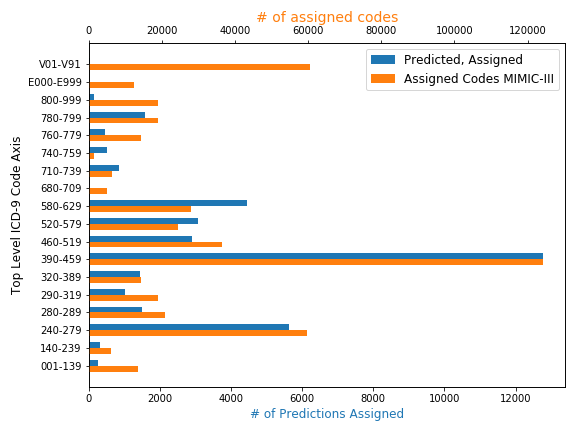}
    \caption{\textbf{Predicted, Assigned Codes} grouped by top-level code group vs total assigned codes}
    \label{fig:predicted_assigned_code_groups}
\end{figure}

We also group the predicted codes into their respective top-level ICD-9 groups in Figure \ref{fig:predicted_assigned_code_groups} and observe that predicted assigned codes display a similar distribution to total assigned codes. We quantify the difference in distributions via the Wasserstein metric or `Earth Movers Distance'\cite{Ramdas2015-ou}. This metric provides a single measure to compare the difference in our 3 datasets distributions when compared with the current assigned code distribution. We compute a small $2.7 \times 10^{-3}$ distance between both distributions, suggesting our method proportionally identifies previously assigned codes from the DD subsection alone. 

\subsubsection{Predicted \& Not Assigned}\label{sec:pred_not_assnd}
This dataset highlights codes that may have been missed from the current assigned codes. 

\begin{figure}
    \centering
    \includegraphics[scale=0.35]{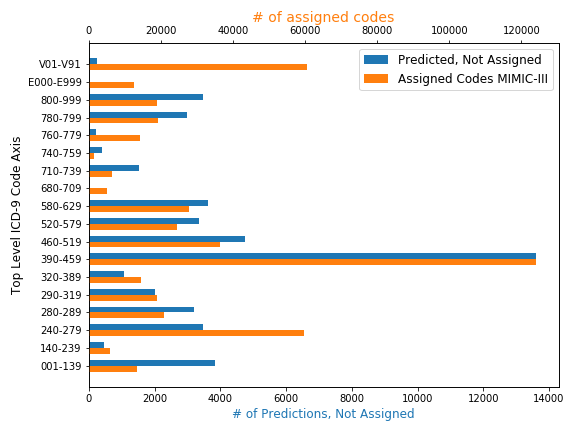}
    \caption{\textbf{Predicted, Not Assigned Codes} grouped by top-level code group vs total assigned codes}
    \label{fig:predicted_not_assigned_code_groups}
\end{figure}

Figure \ref{fig:predicted_not_assigned_code_groups} shows that the distribution of predicted but not assigned codes is minimally different for most codes, supporting our belief that the MIMIC-III assigned codes are not wholly untrustworthy, but are likely under-coded in specific areas.

From this dataset we calculate how many examples of each code that has potentially been missed, or potentially under-coded. For the 10 most frequently assigned codes we see 0-35\%  missing occurrences. We also identify the most frequent code 4019 (Unspecified Essential Hypertension) has 16\% or 3312 potentially missing occurrences. 

To understand if DD subsection length impacts the occurrence of `missed' codes we first calculate a Pearson-Correlation coefficient of $0.17$ for DD subsection line length and counts of assigned codes over all admissions. This suggests a weak positive correlation between admission complexity and number of existing assigned codes. 

In contrast we find a stronger positive correlation of $0.504$ for predicted and not assigned codes and DD subsection line length. This implies that where an episode has a greater number of diagnoses or the complexity of an admission is greater, there is a likelihood to result in more codes being missed during routine collection.

We compute the Wasserstein metric between these two distributions at $1.6 \times 10^{-2}$. This demonstrates a degree of similarity between distributions albeit is 8x further from the Predicted and Assigned dataset distance presented in Section \ref{sec:pred_assned}. We expect to see a larger distance here as we are detecting codes that are indicated in the text but have been missed during routine code assignment.

\subsubsection{Assigned \& Not Predicted}\label{sec:a_np}
We observe that the distribution of assigned and not predicted codes largely mirrors the distribution of total codes assigned in MIMIC-III with a Wasserstein distance of $2.7 \times 10^{-3}$ that is similar to the distance observed in in our Predicted and Assigned Section \ref{sec:pred_assned}) dataset. This suggests that our method is proportionally consistent at not annotating codes that have likely been assigned from elsewhere in the admission, but may also be incorrectly assigned.

\section{Discussion}\label{sec:discussion}
On aggregate, the predicted codes by our MedCAT model suggest that the discharge diagnosis sections listed in 92\% of all discharge notifications are not sufficient for full coding of an episode. Unsurprisingly, this confirms that clinicians infrequently document all codeable diagnoses within the discharge summary. Although, as previously stated, coders are not permitted to make clinical inferences. Therefore, to correctly assign a code, the diagnoses must be present within the documented patient episode within the structured or unstructured data. 

However, the positive correlation between document length and number of predicted codes indicates that missed codes are more prevalent in highly complex cases with many diagnoses. From a coding workflow perspective, coders operate under strict time schedules and are required to code a minimum number of episodes each day. Therefore, it logically follows that the complexity of a case directly correlates to the number of codes missed during routine collection.

Looking at individual code groups we find  240-279 is not predicted proportionally with assigned codes both in P\_A and P\_NA. We explain this as follows. Firstly, DD subsections generally convey clinically important diagnoses for follow-up care. Certain codes such as (250.*) describe diabetes mellitus with specific complications, but the DD subsection will often only describe the diagnoses `DMI' or `DMII'. Secondly, ICU admissions are for individuals with severe illness and therefore are likely to have a high degree of co-morbidity. This is implied by the majority of patients (74\%) are assigned between 4 and 16 codes.

We also observe E000-E999 and V01-V99 codes are disproportionately not predicted. However, this is expected given that both groups are supplementary codes that describe conditions or factors that contribute to an admission but would likely not be relevant for the DD subsection. 

In contrast, we observe a disproportionately large number of predictions for 001-139 (Infectious and Parasitic Diseases). This is primarily driven by 0389 (Unspecified septicemia). A proportion of these predictions may be in error as the specific form of septicemia is likely described in more detail elsewhere in the note and therefore coded as the more specific form. 

\subsection{Method Reproducibility \& Wider Utility}
Inline with the suggestions of \citet{Johnson2017-pr}, the original authors of MIMIC-III, we have attempted to provide the research community all available materials to reproduce and build upon our experiments and method for the development of silver standard datasets. Specifically, we have made the following available as open-source: the SQL scripts to extract the raw data from a replica of the MIMIC-III database, the script required to parse DD subsections, an example script to build a pre-trained MedCAT model, the script required to run MedCAT on the DD subsections, load into the annotator and finally re-run MedCAT and perform experimental analysis alongside outputting the silver standard dataset\footnote{https://github.com/tomolopolis/MIMIC-III-Discharge-Diagnosis-Analysis}.

Given these materials it is possible for researchers to replicate and build upon our method, or directly use the silver standard dataset in future work that investigates automated clinical coding using MIMIC-III. The silver standard dataset clearly marks if each assigned code has been validated or not, or if it is a new code according to our method.

\section{Conclusions \& Future Work}
This work highlighted specific problems with using MIMIC-III as a dataset for training and testing an automated clinical coding system that would limit model performance within a real deployment.

We identified and deterministically extracted the discharge diagnosis (DD) subsections from discharge summaries. We subsequently trained an NER+L model (MedCAT) to extract ICD-9 codes from the DD subsections, comparing the results across the full set of assigned codes. We find our method covers 47\% of all tokens, considering we only take 400 of the $\sim$7k unique codes and perform minimal data cleaning of the DD subsection. We have shown in Section \ref{sec:pred_assned} and \ref{sec:pred_not_assnd} that the MedCAT predicted codes are proportionally inline with assigned codes in MIMIC-III.

Interestingly, we found a $0.504$ positive correlation between DD length and the number of codes predicted by MedCAT, but not assigned in MIMIC-III. This result can be understood by observing that the ICU admissions in MIMIC-III can be extremely complex, with up to 30 clinical codes assigned to a single episode. The DD subsections alone can contain up to 50 line items indicating highly complex cases where codes could easily be missed.

We found that the code group 390-459 (Diseases of the Circulatory System) is both the most assigned group and the group of codes where there are the most missing predictions from our model. Furthermore, codes such as Hypertension (4019), Sepsis and Septicemia (0389, 99591), Gastrointestinal hemorrhage (95789), Chronic Kidney disease (5859), anemia (2859) and Chronic obstructive asthma (49320) are all frequently assigned but are also the highest occurring conditions that appear in the DD diagnosis subsection but are not assigned in the MIMIC-III dataset. This suggests that MIMIC-III exhibits specific cases of undercoding, especially with codes that are frequently occurring in patients but are not likely to be the primary diagnosis for an admission to the ICU. 

As we only use the DD section, there are many codes which likely appear elsewhere in the note that we cannot assign. Although 92\% of discharge summaries contain DD subsections we only match $\sim16\%$ of assigned codes. We suggest this is due to: our NER+L model lacking the ability to identify more synonyms and abbreviations for conditions, the DD subsections lacking enough detail to assign codes and in some occasions, little evidence to suggest a code assignment. Our textual span coverage, presented in Section \ref{sec:pred_coverage} demonstrates that we often cover all available discharge diagnosis, although there is still room for improvement as the majority of the coverage distribution is around the 50\% mark.

For future work we foresee applying the same method to either the entire discharge summary or more specific sections such as `previous medical history' to surface chronic codeable diagnoses that could be validated against the current assigned code set. Researchers would however likely need to address false positive code predictions as clinical coding requires assigned codes to be from current conditions associated with an admission. 

In conclusion, this work has found that frequently assigned codes in MIMIC-III display signs of undercoding up to 35\% for some codes. With this finding we urge researchers to continue to develop automated clinical coding systems using MIMIC-III, but to also consider using our silver standard dataset or build on our method to further improve the dataset.

\section*{Acknowledgments}
RD's work is supported by 1.National Institute for Health Research (NIHR) Biomedical Research Centre at South London and Maudsley NHS Foundation Trust and King’s College London. 2. Health Data Research UK, which is funded by the UK Medical Research Council, Engineering and Physical Sciences Research Council, Economic and Social Research Council, Department of Health and Social Care (England), Chief Scientist Office of the Scottish Government Health and Social Care Directorates, Health and Social Care Research and Development Division (Welsh Government), Public Health Agency (Northern Ireland), British Heart Foundation and Wellcome Trust. 3. The National Institute for Health Research University College London Hospitals Biomedical Research Centre. This paper represents independent research part funded by the National Institute for Health Research (NIHR) Biomedical Research Centre at South London and Maudsley NHS Foundation Trust and King’s College London. The views expressed are those of the author(s) and not necessarily those of the NHS, MRC, NIHR or the Department of Health and Social Care.

\bibliography{anthology,acl2020}
\bibliographystyle{acl_natbib}

\end{document}